\documentclass{llncs2e/llncs}
\usepackage[utf8]{inputenc}

\usepackage[utf8]{inputenc} 
\usepackage[T1]{fontenc}    
\usepackage{hyperref}       
\usepackage{url}            
\usepackage{booktabs}       
\usepackage{amsfonts}       
\usepackage{nicefrac}       
\usepackage{microtype}      
\usepackage{bm}             
\usepackage{graphicx}       
\usepackage{algorithm}      
\usepackage[noend]{algpseudocode} 
\usepackage{caption}        
\usepackage{array}          
\usepackage{booktabs}       
\usepackage{pbox}           
\usepackage{amsmath}        
\usepackage{todonotes}
\usepackage{wrapfig}
\usepackage{aliascnt}
\usepackage{geometry}
\usepackage{natbib}
\title{Distilling a Neural Network Into a Soft Decision Tree}
\author{Nicholas Frosst, Geoffrey Hinton}
\institute{Google Brain Team}
\date{July 2017}

\begin{document}

\maketitle

\begin{abstract}
Deep neural networks have proved to be a very effective way to perform classification tasks. They excel when the input data is high dimensional, the relationship between the input and the output is complicated, and the number of labeled training examples is large \citep{szegedy2015going,wu2016google,jozefowicz2016exploring,graves2013speech}. But it is hard to explain why a learned network makes a particular classification decision on a particular test case. This is due to their reliance on distributed hierarchical representations. If we could take the knowledge acquired by the neural net and express the same knowledge in a model that relies on hierarchical decisions instead, explaining a particular decision would be much easier. We describe a way of using a trained neural net to create a type of soft decision tree that generalizes better than one learned directly from the training data. 

\end{abstract}

\section{Introduction} 

The excellent generalization abilities of deep neural nets depend on their use of distributed representations \citep{lecun2015deep} in their hidden layers, but these representations are hard to understand. For the first hidden layer we can understand what causes an activation of a unit and for the last hidden layer we can understand the effects of activating a unit, but for the other hidden layers it is much harder to understand the causes and effects of a feature activation in terms of variables that are meaningful such as the input and output variables. Also, the units in a hidden layer factor the representation of the input vector into a set of feature activations in such a way that the combined effects of the active features can cause an appropriate distributed representation in the next hidden layer.  This makes it very difficult to understand the functional role of any particular feature activation in isolation since its marginal effect depends on the effects of all the other units in the same layer.

These difficulties are further compounded by the fact that deep neural nets can make reliable decisions by modeling a very large number of weak statistical regularities in the relationship between the inputs and outputs of the training data and there is nothing in the neural network to distinguish the weak regularities that are true properties of the data from the spurious regularities that are created by the sampling peculiarities of the training set.  Faced with all these difficulties, it seems wise to abandon the idea of trying to understand how a deep neural network makes a classification decision by understanding what the individual hidden units do.

By contrast, it is easy to explain how a decision tree makes any particular classification because this depends on a relatively short sequence of decisions and each decision is based directly on the input data.  Decision trees, however, do not usually generalize as well as deep neural nets. Unlike the hidden units in a neural net, a typical node at the lower levels of a decision tree is only used by a very small fraction of the training data so the lower parts of the decision tree tend to overfit unless the size of the training set is exponentially large compared with the depth of the tree.  

In this paper, we propose a novel way of resolving the tension between generalization  and interpretability.  Instead of trying to understand how a deep neural network makes its decisions, we use the deep neural network to train a decision tree that mimics the input-output function discovered by the neural network but works in a completely different way. If there is a large amount of unlabelled data, the neural net can be used to create a much larger labelled data set to train a decision tree, thus overcoming the statistical inefficiency of decision trees. Even if unlabelled data is unavailable, it may be possible to use recent advances in generative modeling \citep{goodfellow2014generative,kingma2013auto} to generate synthetic unlabelled data from a distribution that is close to the data distribution.  Without using unlabelled data, it is still possible to transfer the generalization abilities of the neural net to a decision tree by using a technique called distillation \citep{hinton2015distilling,bucilua2006model} and a type of decision tree that makes soft decisions. 

At test time, we use the decision tree as our model. This may perform slightly worse than the neural network but it will often be much faster and we now have a model whose decisions we can explain and engage with directly.

We start by describing the particular type of decision tree we use. This choice was made to facilitate easy distillation of the knowledge acquired by a deep neural net into a decision tree. 

\section{The Hierarchical Mixture of Bigots} 
We use soft binary decision trees trained with mini-batch gradient descent, where each inner node $i$ has a learned filter ${\bf w}_i$ and a bias $b_i$, and each leaf node $\ell$ has a learned distribution $Q_\ell$. At each inner node, the probability of taking the rightmost branch is:
\begin{equation}
p_i({\bf x}) = \sigma({\bf x}{\bf w}_i + b_i)
\end{equation}
where ${\bf x}$ is the input to the model and $\sigma$ is the sigmoid logistic function. 

This model is a hierarchical mixture of experts \citep{jordan1994hierarchical}, but each expert is a actually a bigot who does not look at the data after training, and therefore always produces the same distribution. The model learns a hierarchy of filters that are used to assign each example to a particular bigot with a particular path probability, and each bigot learns a simple, static distribution over the possible output classes, $k$.

\begin{equation}
    Q^\ell_k = \frac{\exp(\phi^\ell_k)}{\sum_{k'} \exp(\phi^\ell_{k'})},
\end{equation}

where $Q^\ell_\cdot$ denotes the probability distribution at the $\ell^\text{th}$ leaf, and each $\phi^\ell_{\cdot}$ is a learned parameter at that leaf. 

In order to avoid very soft decisions in the tree, we introduced an inverse temperature $\beta$ to the filter activations prior to calculating the sigmoid. Thus the probability of taking the right branch at node $i$ becomes $p_i({\bf x}) = \sigma(\beta({\bf x}{\bf w}_i + b_i))$.

\begin{figure}[t]
\label{treedepth1diargram}
  \centering
    \includegraphics[width=0.60\textwidth]{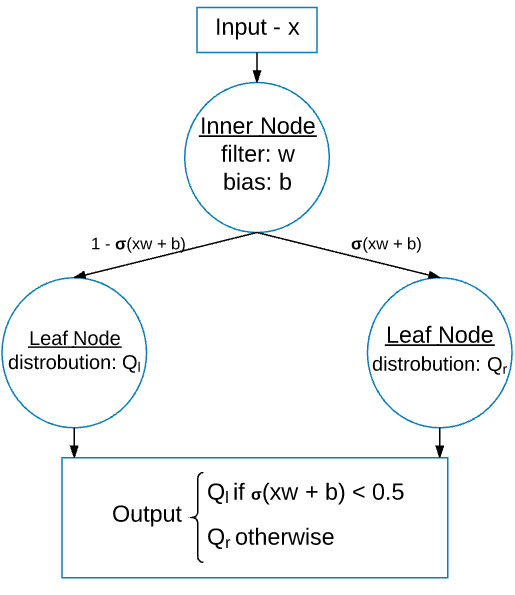}
\caption{This diagram shows a soft binary decision tree with a single inner node and two leaf nodes.}
\end{figure}

This model can be used to give a predictive distribution over classes in two different ways, namely by using the distribution from the leaf with the greatest path probability or averaging the distributions over all the leaves, weighted by their respective path probabilities. If we take the predictive distribution from the leaf with the greatest path probability, the explanation for that prediction is simply the list of all the filters along the path from the route to the leaf together with the binary activation decisions. If we average the leaf distributions weighted by their respective path probabilities, we find that the model achieves marginally better test accuracy, but this leads to an exponential increase in the complexity of the explanation of the model's predictive distribution  on a particular case because it involves the filters at all of the nodes. For this reason, for the remainder of this paper when we refer to the output of the model, we will be referring to the distribution at the leaf with the maximum path probability.

We train the soft decision tree using a loss function that seeks to minimize the cross entropy between each leaf, weighted by its path probability, and the target distribution. For a single training case with input vector ${\bf x}$ and target distribution $T$, the loss is: 

\begin{equation}
\mathrm{L({\bf x})} = - \log\Bigg(\sum_{\ell \in Leaf Nodes} P^\ell({\bf x}) \sum_k T_k \log Q^\ell_k\Bigg) 
\end{equation}
Where $T$ is the target distribution and $P^\ell({\bf x})$ is the probability of arriving at leaf node $\ell$ given the input ${\bf x}$. 

Unlike most decision trees, our soft decision trees use decision boundaries that are not aligned with the axes defined by the components of the input vector. Also, they are trained by first picking the size of the tree and then using mini-batch gradient descent to update all of their parameters simultaneously, rather than the more standard greedy approach that decides the splits one node at a time \citep{friedman2001elements}. 

\section{Regularizers}
To avoid getting stuck at poor solutions during the training, we introduced a penalty term that encouraged each internal node to make equal use of both left and right sub-trees. Without this penalty, the tree tended to get stuck on plateaus in which one or more of the internal nodes always assigned almost all the probability to one of its sub-trees and the gradient of the logistic for this decision was always very close to zero. The penalty is the cross entropy between the desired average distribution ${0.5, 0.5}$ for the two sub-trees and the actual average distribution ${\alpha, (1-\alpha)}$ where $\alpha$ for node $i$ is given by:
\begin{equation}
\alpha_i = \frac{\sum_{{\bf x}} P^i({\bf x}) p_i({\bf x})}{\sum_{{\bf x}} P^i({\bf x})}
\end{equation}
where $P^i({\bf x})$ is the path probability from the root node to node $i$. The penalty summed over all internal nodes is then:
\begin{equation}
C = -\lambda \sum_{i \in InnerNodes} 0.5 \log(\alpha_i) + 0.5 \log(1 - \alpha_i)
\end{equation} 

where $\lambda$ is a hyper-parameter that determines the strength of the penalty and is set prior to training. This penalty was based on the assumption that a tree making fairly equal use of alternative sub-trees would usually be better suited to any particular classification task and in practice it did increase accuracy. However, this assumption is less and less valid as one descends the tree; a penultimate node in the tree may only be responsible for two classes of input, in some non-equal proportion, and penalizing the node for a non-equal split in this case could hurt the accuracy of the model. We found that we achieved better test accuracy results when the strength of the penalty decayed exponentially with the depth $d$ of the node in the tree so that it was proportional to $2^{-d}$.

As one descends the tree, the expected fraction of the data that each node sees in any given training batch decreases exponentially. This means that the computation of the actual probabilities of using the two sub-trees becomes less accurate. To counter this we can maintain an exponentially decaying running average of the actual probabilities with a time window that is exponentially proportional to the depth of the node. We found experimentally that we achieved much better test accuracy by using both the exponential decay in the strength of the penalty with depth and the exponential increase in the temporal scale of the window used to compute the running average.

\section{MNIST Results} 
The number of total parameters at which our soft decision trees start to overfit is typically less than the number of total parameters at which a multi-layer neural network starts to overfit. This is because the lower nodes of the decision tree only receive a very small fraction of the training data.  


This is reflected in performance on MNIST. With a soft decision tree of depth 8 we were able to achieve a test accuracy of at most 94.45\% when training on the true targets. A neural net with two convolutional hidden layers and a penultimate fully connected layer achieved a much better test accuracy of 99.21\%. We were then able to use the accuracy of the neural net to make a much better soft decision tree by training with soft targets that were a composite of the true labels and the predictions of the neural network. The soft decision tree trained in this way 
achieved a test accuracy of 96.76\% which is about halfway between the neural net and the soft decision tree trained directly on the data.

\section{Explaining how a soft decision tree makes a classification} 
\begin{figure}[t]
\label{treedepth4}
  \centering
    \includegraphics[width=\textwidth]{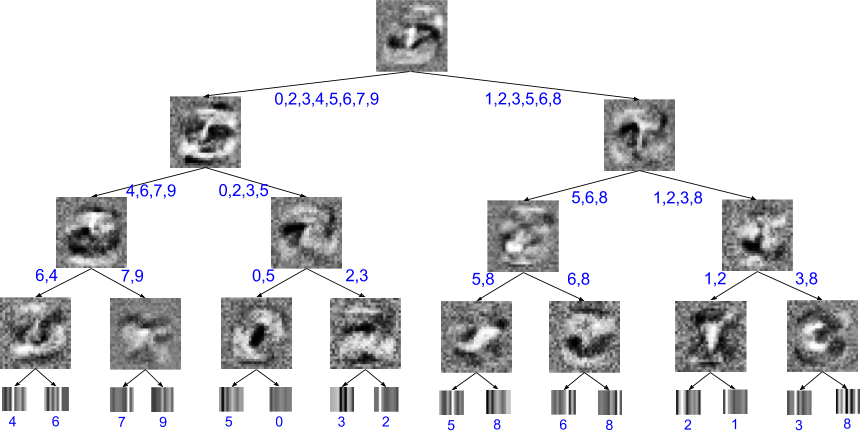}
\caption{This is a visualization of a soft decision tree of depth 4 trained on MNIST. The images at the inner nodes are the learned filters, and the images at the leaves are visualizations of the learned probability distribution over classes. The final most likely classification at each leaf, as well as the likely classifications at each edge are annotated. If we take for example the right most internal node, we can see that at that level in the tree the potential classifications are only 3 or 8, thus the learned filter is simply learning to distinguish between those two digits. The result is a filter that looks for the presence of two areas that would join the ends of the 3 to make an 8.}
\end{figure}
The main motivation behind this work was to create a model whose behavior is easy to explain; in order to fully understand why a particular example was given a particular classification, one can simply examine all the learned filters along the path between the root and the classification's leaf node. 
The crux of this model is that it does not rely on hierarchical features, it relies on hierarchical decisions instead. The hierarchical features of a traditional neural network allow it to learn robust and novel representations of the input space, but past a single level or two, they become extremely difficult to engage with. Some current attempts at explanations for neural networks rely on the use of gradient descent to find an input that particularly excites a given neuron \citep{simonyan2013deep,erhan2009visualizing}, but this results is a single point on a manifold of inputs, meaning that other inputs could yield the same pattern of neural excitement, and so it does not reflect the entire manifold. Ribeiro et al. propose a strategy which relies on fitting some explainable model which "acts over absence/presence of interpretable components" to the behavior of a deep neural net around some area of interest in the input space \citep{DBLP:journals/corr/RibeiroSG16}. This is accomplished by sampling from the input space and querying the model around the area of interest and then fitting an explainable model to the output of the model. This avoids the problem of attempting to explain a particular output by visualizing a single point on a manifold but introduces the problem of necessitating a new explainable model for every area of interest in the input space, and attempting to explain changes in the model's behavior by first order changes in a discretized interpretation of the input space. By relying on hierarchical decisions instead of hierarchical features we side-step these problems, as each decision is made at a level of abstraction that the reader can engage with directly.

\section{Other Data Sets and Results} 
We tried this model on several other data sets, but focused on spatial input for the sake of visualization. By first training a neural net and then using it to provide soft targets for training a soft decision tree, with a tree of depth 8 we were able to achieve a test accuracy of 80.60\% on the Connect4 dataset \citep{Lichman:2013} comprised of board states of the popular child's game connect 4 as input, and the final outcome of the game (player 1 win, player 2 win, or tie) as the target value. Without distilling from a neural net, the best test accuracy we achieved was 78.63\%. Other decision trees trained with gradient descent have been applied to this dataset \citep{norouzi2015efficient} but were only able to achieve a maximum test accuracy of 76.50\% at the equivalent depth of 8 and 77.45\% at a depth of 20. This provides an interesting example of the utility of an explainable model - by examining the learned filters of the soft decision tree we are able to learn something about the nature of the game. From examining the first learned filter we can see that the game can be split into two distinct sub types of games - games where the players have placed pieces on the edges of the board, and games where the players have placed pieces in the center of the board. These two sub games progress in sufficiently different manners that it was beneficial for the decision tree to split them at the root.

\begin{figure}[h]
\label{connect4vis}.
  \centering
    \includegraphics[width=0.70\textwidth]{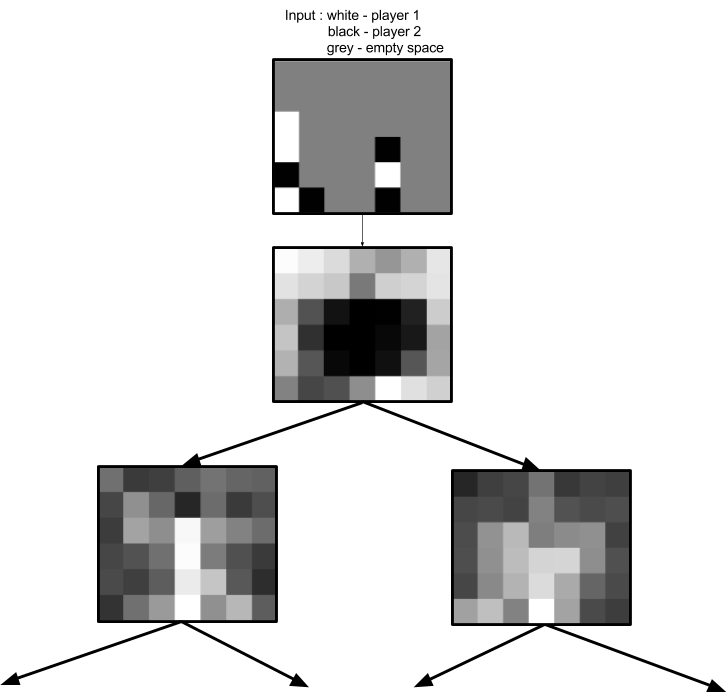}
\caption{This is a visualization of the first 2 layers of a soft decision tree trained on the Connect4 data set. From examining the learned filters we can see that the game can be split into two distinct sub types of games - games where the players have placed pieces on the edges of the board, and games where the players have placed pieces in the center of the board.}
\end{figure}

We also ran our model on a non spatial dataset, namely the Letter dataset \citep{Lichman:2013}, which is comprised of primitive numerical attributes of capital english characters. We were able to achieve a test accuracy of 78.0\% with a tree of depth 9 trained on the raw training data, and a test accuracy of 81.0\% when we distilled from an ensemble of neural nets that had a 95.9\% test accuracy.

\section{Conclusion}
We have described a method for using a trained neural net to create a more explicable model in the form of a soft decision tree which is trained by stochastic gradient descent using the predictions of the neural net to give more informative targets. The soft decision tree uses learned filters to make hierarchical decisions based on an input example and ultimately select a particular static probability distribution over classes as its output. This soft decision tree generalizes better than one trained on the data directly, but performs worse than the neural net which was used to provide the soft targets for training it. 
So if it is essential to be able to explain why a model classifies a particular test case in a particular way, we can use a soft decision tree, but we can still gain some of the benefits of deep neural networks by using them to improve the training of this explicable model.

\bibliographystyle{unsrtnat}
\bibliography{neuraldecisiontree}

\end{document}